%% file: main.tex
\documentclass[fleqn,10pt]{wlscirep}
\usepackage[utf8]{inputenc}
\usepackage[T1]{fontenc}
\usepackage[ruled,vlined]{algorithm2e}
\usepackage{xcolor}
\usepackage{multirow}
\author[1]{Gorkem Can Ates}
\author[2]{Yu Xin}
\author[3]{Kuang Gong}
\author[1,2,4,*]{Wei Shao}

\affil[1]{Department of Medicine, University of Florida, Gainesville, FL, 32611, USA}
\affil[2]{Department of Electrical and Computer Engineering, University of Florida, Gainesville, FL, 32611, USA}
\affil[3]{Department of Biomedical Engineering, University of Florida, Gainesville, FL, 32611, USA}
\affil[4]{Intelligent Clinical Care Center, University of Florida, Gainesville, FL, 32611, USA}
\affil[*]{Corresponding author. E-mail address: weishao@ufl.edu}

\newcommand{\dc}{DCFormer}

\begin{document}
\title{DCFormer: Efficient 3D Vision-Language Modeling with Decomposed Convolutions} %DCFormer

\input{1-abstract}

\flushbottom
\maketitle

\input{2-introduction}
\thispagestyle{empty}

\input{3-results}

\input{4-discussion}

\input{5-methods}

\newpage

\bibliography{ref}

\section*{Acknowledgements}

This work was supported by the Department of Medicine and the Intelligent Clinical Care Center at the University of Florida College of Medicine. The authors express their sincere gratitude to the NVIDIA AI Technology Center at the University of Florida for their invaluable feedback, technical guidance, and support throughout this project.

\section*{Author Contributions}

G.C.A. conducted the study design, software development, experimental setup, data analysis, visualization, and manuscript writing. Y.X. contributed to the experimental setup, manuscript writing, and discussions. K.G. provided valuable insights and contributed to discussions throughout the study. W.S. supervised and conceptualized the project, guided methodological development and experimental setup, and contributed to writing the original manuscript. All authors reviewed and approved the final manuscript.

\section*{Additional Information}

%\textbf{Accession codes} (where applicable); 

\textbf{Competing interests:} The authors have no conflict of interest to declare. 

% The corresponding author is responsible for submitting a \href{http://www.nature.com/srep/policies/index.html#competing}{competing interests statement} on behalf of all authors of the paper. This statement must be included in the submitted article file.

%\begin{figure}[ht]
%\centering
%\includegraphics[width=\linewidth]{stream}
%\caption{Legend (350 words max). Example legend text.}
%\label{fig:stream}
%\end{figure}

\end{document}

%% file: 1-abstract.tex
\begin{abstract}
Vision-language models (VLMs) have been widely applied to 2D medical image analysis due to their ability to align visual and textual representations. However, extending VLMs to 3D imaging remains computationally challenging. Existing 3D VLMs often rely on Vision Transformers (ViTs), which are computationally expensive due to the quadratic complexity of self-attention, or on 3D convolutions, which require large numbers of parameters and FLOPs as kernel size increases. We introduce DCFormer, an efficient 3D image encoder that factorizes 3D convolutions into three parallel 1D convolutions along the depth, height, and width dimensions. This design preserves spatial information while significantly reducing computational cost. Integrated into a CLIP-based vision-language framework, DCFormer is trained and evaluated on CT-RATE, a dataset of 50,188 paired 3D chest CT volumes and radiology reports. In zero-shot and fine-tuned detection of 18 pathologies, as well as in image–text retrieval tasks, DCFormer consistently outperforms state-of-the-art 3D vision encoders, including CT-ViT, ViT, ConvNeXt, PoolFormer, and TransUNet. These results highlight DCFormer’s potential for scalable, clinically deployable 3D medical VLMs. Our code is available at: \url{https://github.com/mirthAI/DCFormer}.

\end{abstract}

%% file: 2-introduction.tex
\section*{Introduction}
Deep learning has revolutionized medical imaging by enabling automated disease diagnosis, accurate prognosis, and personalized treatment planning \cite{litjens2017survey, esteva2021deep, shen2017deep, fink2020potential}. In particular, Convolutional Neural Networks (CNNs) \cite{fukushima1980neocognitron, lecun1989handwritten} and Vision Transformers (ViTs) \cite{dosovitskiy2020image} have driven state-of-the-art performance in many supervised tasks (e.g., lesion detection and organ segmentation \cite{ronneberger2015u, chen2021transunet, cao2022swin, wang2022uctransnet, ates2023dual}). However, these supervised methods require training on large-scale manually annotated datasets and often fail to generalize across different imaging modalities and tasks. This limitation has motivated the development of more scalable and generalizable learning paradigms.

To reduce the dependency on extensive data labeling, self-supervised learning (SSL) methods such as masked autoencoders \cite{he2022masked} have been developed to extract informative representations from unlabeled images \cite{tang2022self, chen2022scaling}. While effective, these SSL methods still typically require subsequent supervised fine-tuning on small labeled datasets. Vision-language models (VLMs) such as CLIP (Contrastive Language-Image Pretraining) \cite{radford2021learning} offer a promising alternative by aligning visual and textual representations in a shared latent space, enabling zero-shot capabilities. Initially developed for natural image domains \cite{lin2022frozen, thengane2022clip}, CLIP has since been adapted for 2D medical imaging tasks, including zero-shot classification, prompt-based segmentation, image-text retrieval, radiology report generation, and visual question answering \cite{koleilat2024medclip, lu2024multimodal, tiu2022expert, endo2021retrieval, hu2022x, eslami2023pubmedclip, huang2023visual}.

Despite the success of VLMs in 2D, their application to 3D imaging remains underexplored due to several challenges. One major challenge is the scarcity of large-scale public datasets of 3D image volumes paired with text reports. Recently, Hamamci et al. \cite{hamamci2024foundation} addressed this by releasing CT-RATE, a dataset consisting of 50,188 reconstructed 3D chest computed tomography (CT) volumes (from 25,692 scans of 21,304 patients) with corresponding radiology reports. They also introduced CT-CLIP, a 3D VLM for chest CT that uses a vision transformer encoder (CT-ViT \cite{hamamci2025generatect}) with a two-stage spatial and temporal transformer architecture. Although CT-CLIP achieved promising results in zero-shot detection and image-text retrieval tasks, its reliance on self-attention for the image encoder leads to significant computational overhead. Additionally, CT-CLIP uses a ViT-like patch tokenization strategy that downsamples each volume using a fixed patch size of 20. This strategy causes a loss of fine-grained information, as it fails to extract hierarchical representations — an essential element in medical imaging. Prior work has shown that capturing multi-scale, hierarchical features is crucial for understanding complex spatial and contextual information in medical images \cite{chen2021transunet, cao2022swin, hatamizadeh2021swin}. Nonetheless, CT-CLIP marks a significant step toward bringing VLMs to 3D medical imaging.

\input{figures/conv_flop_merge}
Another barrier to 3D VLMs is the computational complexity of processing volumetric data (e.g., CT scans). High-resolution 3D image volumes impose heavy computational demands on standard ViT-based or 3D CNN-based encoders. While ViTs are effective at capturing global relationships, their computational complexity scales quadratically with input size, making them inefficient for 3D volumes. Similarly, 3D convolutions are well-suited for local feature extraction, but they can become prohibitively expensive as convolutional kernel size grows beyond 3. To mitigate this, depthwise convolutions \cite{chollet2017xception} have been explored as an alternative, reducing both parameter count and computational cost. However, they can still be inefficient for 3D imaging when large kernels and high output channel dimensions are required. Figure \ref{fig:convflop_merge} compares the parameters and FLOPs of 2D and 3D standard versus depthwise convolutions, showing that standard 3D convolutions demand significantly more resources than their 2D counterparts. 
Although depthwise convolutions improve efficiency, their computational cost scales cubically with kernel size, posing challenges for 3D medical imaging where large receptive fields are needed to capture complex anatomical structures. Additionally, stacking many such layers in a deep network compounds computational burden. 

To address these challenges of existing 3D VLMs, we propose \dc, a family of 3D vision-language encoder architectures for medical images. \dc\ replaces both self-attention and standard 3D convolutions with a decomposed convolution strategy that significantly reduces computational complexity while preserving the ability to extract both local and global spatial features. In essence, each 3D convolution in DCFormer is factorized into three 1D convolutions applied along the height, width, and depth dimensions (Figure \ref{fig:deconvnext}). This factorization significantly reduces the number of  parameters and FLOPs required (Figure \ref{fig:compdecomp}), achieving a much better balance of efficiency and performance. We integrate \dc\ as the vision encoder into a CLIP framework to enable efficient image-text representation learning on 3D imaging data. In our experiments on the CT-RATE dataset, \dc\ achieves competitive or better performance in zero-shot pathology detection and image-text retrieval tasks, while significantly reducing computational costs compared to existing state-of-the-art methods. To our knowledge, \dc\ is the first 3D vision encoder to use decomposed convolutions to improve vision-language modeling. We anticipate that \dc\ can serve as a stepping stone toward scalable, efficient image encoders for vision-language understanding, and hope it will inspire further research into lightweight, high-performance architectures for clinical applications.
\input{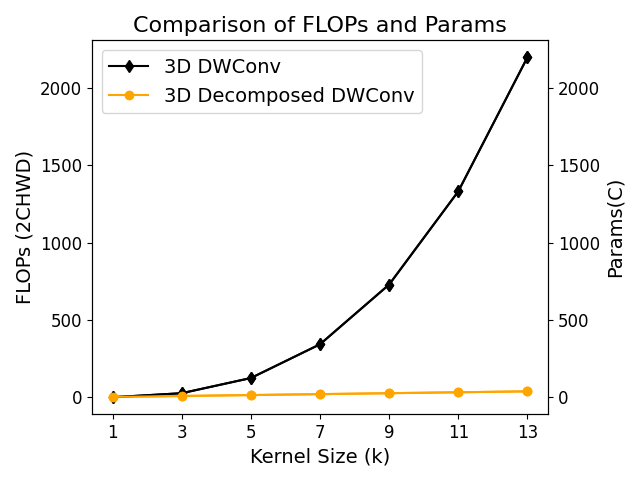}

%% file: figures/conv_flop_merge.tex
\begin{figure*}[t]
\centering
\includegraphics[width=14cm]{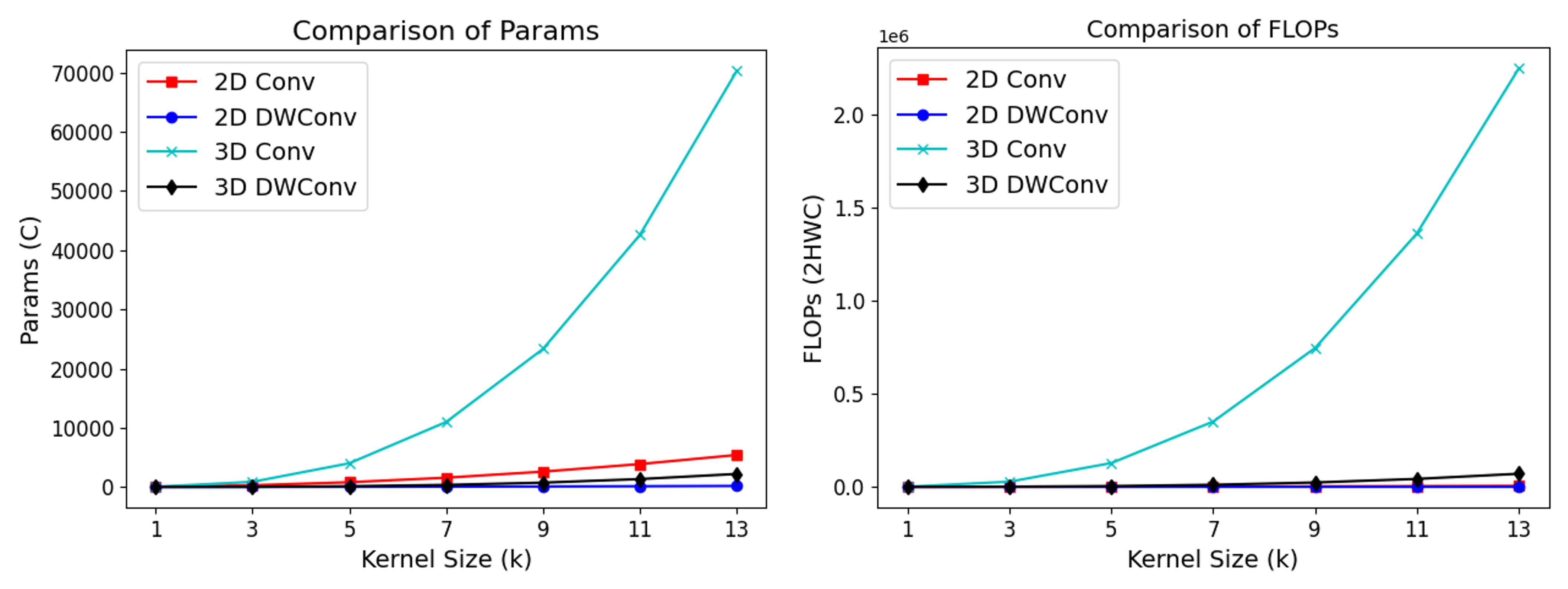}\\
\caption{Parameter count and computational cost (FLOPs) comparison for 2D and 3D standard and depthwise convolutions across kernel sizes. For simplicity, the number of output channels is fixed at $C=32$, and the depth dimension is set to $D=32$.}
\label{fig:convflop_merge} 
\end{figure*} 

%% file: figures/compdecomp.tex
\begin{figure*}[!ht]
\centering
\includegraphics[width=7cm]{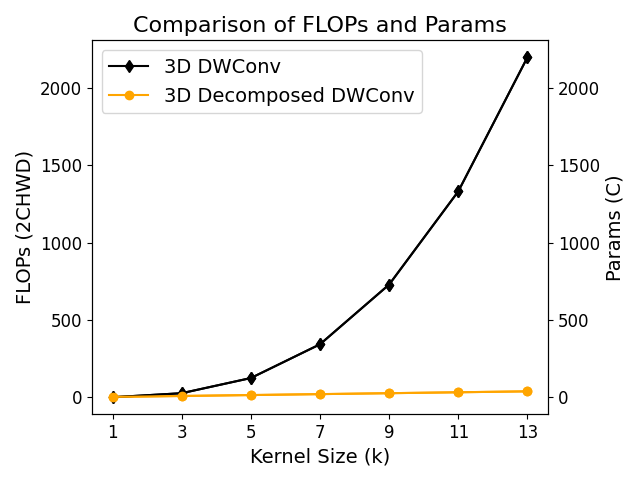}\\
\caption{Parameter count and computational cost (FLOPs) comparison for standard 3D depthwise convolution and decomposed depthwise convolution.}
\label{fig:compdecomp} 
\end{figure*}

%% file: 3-results.tex
\section*{Results}
We trained a CLIP-based vision-language framework (see Methods: Developing a CLIP framework using \dc\ ) using the CT-RATE dataset, which consists of 50,188 non-contrast 3D chest CT volumes paired with corresponding free-text radiology reports. For the text encoder, we used a domain-specific BERT model (CXR-BERT \cite{boecking2022making}). For the image encoder, we evaluated different variants of our proposed \dc\ architecture (nano, naïve, tiny) against several state-of-the-art 3D vision encoders, including ViT3D \cite{dosovitskiy2020image}, CT-ViT \cite{hamamci2025generatect, hamamci2024foundation}, TransUNet \cite{chen2021transunet}, ConvNeXt \cite{liu2022convnet}, and PoolFormer \cite{yu2022metaformer}. We focused on three main tasks: zero-shot abnormality detection, fine-tuned abnormality detection, and image–text retrieval.

\subsection*{Zero-Shot Multi-Abnormality Detection}
Once trained to align images and text in a shared latent space, our CLIP model can perform zero-shot detection of multiple pathologies in a chest CT volume by using each pathology name as a text prompt (Figure \ref{fig:clip}b)). We adopt an approach similar to CheXzero\cite{tiu2022expert} and CT-CLIP\cite{hamamci2024foundation}: for each of the 18 candidate abnormalities, we provide a positive ('\{\textit{Pathology}\} is present.') prompt and a negative ('\{\textit{Pathology}\} is not present.') prompt. We compute the cosine similarity between between the CT image embedding and each of the text prompt embeddings.
These similarities are used to compute the normalized probability of each abnormality being present.
We evaluate model performance using accuracy, F1 score, precision, and recall.
 \input{tables/results3}

Table \ref{results} compares the zero-shot performance and computational efficiency of multiple models on the CT-RATE dataset (image volume size 512×512×256). DCFormer provides a family of models with configurations ranging from nano to tiny variants. Across all configurations, the lightweight variants of DCFormer consistently achieve superior results while reducing computational overhead. For instance, the nano variant of DCFormer achieves higher F1 score, precision, and recall than the corresponding nano variants of ConvNeXt and PoolFormer. Similarly, the naïve variant of DCFormer achieves a higher F1 score than the naïve variants of ConvNeXt, PoolFormer, and TransUNet. Notably, DCFormer achieves these results with significantly improved computational efficiency. For example, the naïve variant of DCFormer delivers a 44.5\% F1 score—outperforming CT-ViT as well as the tiny variants of ConvNeXt, PoolFormer, and TransUNet—while using substantially fewer parameters and FLOPs: 5.85 million parameters and 49.48 GFLOPs, compared to CT-ViT’s 101.1 million parameters and 160.5 GFLOPs. DCFormer’s ability to outperform state-of-the-art models with far fewer computational resources underscores its potential for applications demanding both high performance and efficiency.

 \subsection*{Fine-Tuned Multi-Abnormality Detection}
  \input{tables/results_finetune_2}
To further validate \dc\, we fine-tuned each model by adding a single linear classification layer on top of the frozen image encoder, and trained on labeled data for the same 18 chest abnormalities (Figure \ref{fig:clip}c). 
We used a Binary Cross-Entropy (BCE) loss to allow independent prediction of each abnormality (i.e., treating it as a multi-label classification rather than multi-class). Table \ref{results_finetune} summarizes the fine-tuning results on the test set. As expected, fine-tuning improves the F1 scores of all models compared to their zero-shot performance. Notably, DCFormer variants continue to outperform all other architectures when fine-tuned.
The DCFormer-tiny model achieves the highest F1 score at 48.6\%, surpassing the best competing model (CTViT, 46.4\%). DCFormer-naïve also achieves a strong F1 score of 47.2\%, outperforming all other fine-tuned models.  These gains highlight DCFormer's efficient and effective feature extraction when adapted to specific tasks.

 \subsection*{Image-Text Retrieval }
Finally, we evaluated DCFormer on bidirectional image-text retrieval tasks. We computed cosine similarity scores between each CT volume embedding and each report embedding, and evaluated retrieval performance using Recall@k. As shown in Table \ref{results_ret}, DCFormer delivers strong performance in both image-to-text (IR) and text-to-image (TR) tasks. The DCFormer tiny variant achieves the highest Recall@50, with 11.31\% for IR and 12.37\% for TR. The naïve variant of DCFormer significantly outperforms the nano variant, highlighting the effective scaling capability of our model. These results demonstrate DCFormer's ability to capture semantic relationships between visual and textual data. We attribute the strong retrieval results to DCFormer's use of large-kernel decomposed convolutions, which provide broad spatial context while maintaining efficiency.
 
  \input{tables/results_retrieval}

%% file: tables/results3.tex
\begin{table*}[!ht]
\scriptsize
\renewcommand\arraystretch{1.4}
\begin{center}
\caption{\label{results} Zero-shot performance of models trained on CT-RATE dataset at the resolution of 512x512x256.}
\begin{tabular}{c|lll|cccc}
    \toprule
    \textbf{Model} & \textbf{Variant} & \textbf{Params (M)} & \textbf{GFLOPS} & \textbf{Accuracy (\%)} & \textbf{F1 Score (\%)} & \textbf{Precision (\%)} & \textbf{Recall (\%)} \\ \midrule
    \multirow{3}{*}{DCFormer} & nano & 0.92 & 34.21 & 60.4 & 41.9 & 27.2 & 62.8 \\
                              & naïve & 5.85 & 49.48 & 63.1 & 44.5 & 29.5 & 65.5 \\
                              & tiny & 15.1 & 168.2 & 62.0 & 46.3 & 29.7 & 70.1 \\ \midrule
    \multirow{3}{*}{ConvNeXt \cite{liu2022convnet}} & nano & 3.19 & 31.92 & 62.2 & 39.4 & 26.7 & 55.1 \\
                              & naïve & 15.63 & 96.84 & 60.7 & 42.4 & 27.7 & 63.8 \\
                              & tiny & 31.59 & 156.31 & 62.5 & 42.1 & 28.2 & 60.1 \\ \midrule
    \multirow{3}{*}{PoolFormer \cite{yu2022metaformer}} & nano & 2.79 & 27.14 & 60.2 & 37.0 & 24.8 & 52.3 \\
                              & naïve & 11.31 & 63.75 & 60.1 & 39.1 & 25.7 & 56.8 \\
                              & tiny & 20.68 & 117.46 & 61.8 & 38.3 & 26.0 & 53.5 \\ \midrule
    \multirow{2}{*}{TransUNet \cite{chen2021transunet}} & naïve & 12.48 & 118.9 & 58.6 & 41.4 & 26.5 & 56.0 \\
                              & tiny & 23.93 & 207.5 & 61.5 & 35.8 & 24.7 & 48.7 \\ \midrule
    \multirow{2}{*}{ViT \cite{dosovitskiy2020image}} & naïve & 11.10 & 39.05 & 55.0 & 42.5 & 25.8 & 71.5 \\
                              & tiny & 26.34 & 86.43 & 61.0 & 43.2 & 28.0 & 64.8 \\ \midrule
    \textbf{CTViT \cite{hamamci2024foundation}} & - & 101.1 & 160.5 & 62.9 & 44.3 & 29.3 & 65.7 \\ 
    \bottomrule
\end{tabular}
\end{center}
\end{table*}

%% file: tables/results_finetune_2.tex
\begin{table*}[t]
\scriptsize
\renewcommand\arraystretch{1.4}
\begin{center}
\caption{\label{results_finetune} Fine-tuning performance of models trained on CT-RATE dataset at the resolution of 512x512x256.}

\begin{tabular}{c|lll|cccc}
    \toprule
    \textbf{Model} & \textbf{Variant} & \textbf{Params (M)} & \textbf{GFLOPS} & \textbf{Accuracy (\%)} & \textbf{F1 Score (\%)} & \textbf{Precision (\%)} & \textbf{Recall (\%)} \\ \midrule
    \multirow{3}{*}{DCFormer} & nano & 0.92 & 34.21 & 60.1 & 45.0 & 28.6 & 69.6 \\
                              & naïve & 5.85 & 49.48 & 64.2 & 47.2 & 31.0 & 70.2 \\
                              & tiny & 15.1 & 168.2 & 64.1 & 48.6 & 31.4 & 72.9 \\ \midrule
    \multirow{3}{*}{ConvNeXt \cite{liu2022convnet}} & nano & 3.19 & 31.92 & 56.3 & 43.5 & 26.2 & 70.1 \\
                              & naïve & 15.63 & 96.84 & 59.0 & 44.2 & 27.2 & 67.6 \\
                              & tiny & 31.59 & 156.31 & 60.4 & 44.8 & 28.2 & 68.2 \\ \midrule
    \multirow{3}{*}{PoolFormer \cite{yu2022metaformer}} & nano & 2.79 & 27.14 & 52.6 & 41.2 & 24.1 & 67.6 \\
                              & naïve & 11.31 & 63.75 & 53.9 & 41.9 & 24.8 & 68.7 \\
                              & tiny & 20.68 & 117.46 & 54.2 & 42.2 & 25.0 & 68.8 \\ \midrule
    \multirow{2}{*}{TransUNet \cite{chen2021transunet}} & naïve & 12.48 & 118.9 & 60.2 & 44.6 & 29.4 & 70.1 \\
                              & tiny & 23.93 & 207.5 & 50.2 & 39.4 & 22.7 & 66.1 \\ \midrule
    \multirow{2}{*}{ViT \cite{dosovitskiy2020image}} & naïve & 11.10 & 39.05 & 56.2 & 43.1 & 26.1 & 69.8 \\
                              & tiny & 26.34 & 86.43 & 59.4 & 44.9 & 28.0 & 70.8 \\ \midrule
    \textbf{CTViT \cite{hamamci2024foundation}} & - & 101.1 & 160.5 & 61.8 & 46.4 & 29.4 & 70.1 \\ \bottomrule
\end{tabular}
\end{center}
\end{table*}

%% file: tables/results_retrieval.tex
\begin{table*}[!ht]
\scriptsize
\renewcommand\arraystretch{1.4}
\begin{center}
\caption{\label{results_ret} Image-text retrieval performance of models trained on CT-RATE dataset at the resolution of 512x512x256.}
\begin{tabular}{c|l|ccc|ccc}
\toprule
\multirow{2}{*}{\textbf{Models}}     & \multirow{2}{*}{\textbf{Variant}}        & \multicolumn{3}{c|}{\textbf{IR}}                                          & \multicolumn{3}{c}{\textbf{TR}}                                            \\
\cline{3-8}
 &  & \textbf{R@1}    & \textbf{R@10}   & \textbf{R@50}   & \textbf{R@1}    & \textbf{R@10}   & \textbf{R@50}   \\ \midrule
\multirow{3}{*}{DCFormer}    & nano                         & 0.0001 & 0.0135 & 0.0617 & 0.0001 & 0.0135 & 0.0707 \\ 
                             & naive                        & 0.0026 & 0.0220 & 0.0954 & 0.0016 & 0.0214 & 0.0971 \\
                             & tiny                         & 0.0053 & 0.0296 & 0.1131 & 0.0039 & 0.0280 & 0.1237 \\ \midrule
\multirow{3}{*}{ConvNeXt\cite{liu2022convnet}}    & nano                         & 0.0013 & 0.0102 & 0.0562 & 0.0029 & 0.0135 & 0.0599 \\ 
                             & naive                        & 0.0016 & 0.0161 & 0.0780 & 0.0023 & 0.0214 & 0.0740 \\
                             & tiny                         & 0.0026 & 0.0135 & 0.0694 & 0.0010 & 0.0171 & 0.0796 \\ \midrule
\multirow{3}{*}{PoolFormer\cite{yu2022metaformer}}  & nano                         & 0.0016 & 0.0095 & 0.0480 & 0.0023 & 0.0164 & 0.0592 \\ 
                             & naive                        & 0.0013 & 0.0134 & 0.0569 & 0.0026 & 0.0151 & 0.0635 \\ 
                             & tiny                         & 0.0007 & 0.0171 & 0.0638 & 0.0016 & 0.0115 & 0.0648 \\ \midrule
\multirow{2}{*}{TransUNet\cite{chen2021transunet}}   & naive                        & 0.0023 & 0.0236 & 0.1000 & 0.0033 & 0.0269 & 0.1105 \\
                             & tiny                         & 0.0016 & 0.0131 & 0.0589 & 0.0026 & 0.0151 & 0.0651 \\ \midrule
\multirow{2}{*}{ViT\cite{dosovitskiy2020image}}         & naive                        & 0.0016 & 0.0112 & 0.0500 & 0.0007 & 0.0109 & 0.0563 \\ 
                             & tiny                         & 0.0010 & 0.0167 & 0.0628 & 0.0013 & 0.0148 & 0.0737 \\ \midrule
\textbf{CTViT}\cite{hamamci2024foundation}                        & -                            & 0.0046 & 0.0293 & 0.1096 & 0.0023 & 0.0260 & 0.0921 \\ \bottomrule
\end{tabular}
\end{center}

\end{table*}

%% file: 4-discussion.tex
\section*{Discussion}
In this paper, we introduced DCFormer, a novel 3D vision encoder designed to efficiently process volumetric medical images within vision–language models. DCFormer addresses key computational challenges in 3D image analysis, where conventional models struggle due to the quadratic complexity of vision transformers and the computational burden of large-kernel 3D convolutions. By leveraging decomposed 3D convolutions as a token mixing mechanism, DCFormer drastically reduces parameter count while preserving strong spatial feature extraction capabilities. 

We evaluated DCFormer on CT-RATE, a large-scale dataset comprising 50,188 3D chest CT volumes paired with corresponding radiology reports, benchmarking its performance against several state-of-the-art 3D vision encoders, including ViT, ConvNeXt, PoolFormer, TransUNet, and CT-ViT. Across tasks such as multi-abnormality detection (in both zero-shot and fine-tuned settings) and image–text retrieval, DCFormer consistently outperformed all competing models. This performance is directly attributable to its decomposed convolution architecture, which captures multi-scale spatial features without the computational burden of self-attention or full 3D convolutions. Unlike CT-ViT and other resource-intensive models, DCFormer achieves a favorable balance between accuracy and efficiency, making it a practical and scalable solution for volumetric medical imaging.

Importantly, DCFormer’s low computational cost and fast inference time make it well suited for clinical applications requiring real-time responsiveness—such as automated triage, abnormality flagging, or preliminary report generation. With only 5.85 million parameters, the DCFormer-naïve variant can run on standard hospital hardware without requiring specialized infrastructure, significantly lowering the barrier to deploying advanced AI in routine workflows. We envision DCFormer as a foundational component in next-generation clinical decision support systems, helping democratize access to multimodal AI and accelerating the adoption of vision–language tools in healthcare.

Beyond the CLIP framework, DCFormer has been independently validated as a versatile vision backbone in the recently proposed Med3DVLM architecture \cite{xin2025med3dvlm}. In that work, DCFormer was combined with SigLIP \cite{zhai2023sigmoid} and incorporated into the LLaVA \cite{liu2024visual} framework to enable instruction-tuned multimodal understanding over 3D medical volumes and free-text prompts. The resulting system demonstrated competitive or superior performance in 3D report generation, zero-shot retrieval, and visual question answering tasks. This broader validation highlights DCFormer’s generalizability across different vision-language learning paradigms and tasks, reinforcing its potential as a robust foundation for multimodal medical AI.

While DCFormer demonstrates strong potential, several avenues for future work remain.  First, more sophisticated prompt engineering techniques could further improve zero-shot performance by better aligning the visual and textual embeddings for each clinical query. Second, incorporating additional and more diverse training data – including other imaging modalities like MRI and PET  – could improve the model’s generalization and validate its effectiveness across different 3D imaging contexts. Finally, integrating DCFormer with emerging multimodal frameworks (such as LLaVA \cite{liu2024visual} for vision–language question answering or LISA \cite{lai2024lisa} for segmentation) could broaden its application to tasks like interactive diagnosis and automated 3D image interpretation.

%% file: 5-methods.tex
\section*{Methods}
%We propose DCFormer, a novel token mixer to efficiently process 3D CT images in the context of a CLIP-based framework. DCFormer addresses the computational challenges inherit in volumetric 3D CT scans by decomposing the computationally expensive 3D convolution into several 1D operations across all spatial dimensions. DCFormer utilizes MetaFormer\cite{yu2022metaformer, yu2023metaformer} as the backbone block design and uses decomposed convolutions as the token mixer for feature extraction. It also enjoys a hierarchical structure which ultimately enhances feature extraction capacity for 3D CT images. Finally, the proposed DCFormer-based image encoder is integrated into a CLIP framework to achieve text-image alignment. As we will show later, such a decomposition strategy significantly reduces parameter count and computational costs while preserving high zero-shot and fine-tuning performance. 

%\subsection*{DCFormer}

Processing 3D images presents significant computational challenges due to their high resolution and volumetric complexity. The computational complexity of standard 3D convolutions scales cubically with kernel size, and self-attention mechanisms further amplify the burden by scaling quadratically with image size. To address these challenges, we developed DCFormer, a hybrid architecture that decomposes 3D convolutions into multiple 1D components, significantly reducing computational overhead. DCFormer adopts the MetaFormer paradigm \cite{yu2022metaformer, yu2023metaformer} as its backbone and employs decomposed convolutions as the token mixer for feature extraction. Its hierarchical design (described below) further enhances feature representation in 3D images. Finally, the DCFormer-based image encoder is integrated into a CLIP framework for image-text alignment. As demonstrated in our results, this decomposition strategy substantially reduces parameter count and computational cost while maintaining high performance.

\subsection*{Formulation of the DCFormer block} %% may need some updates attention part

Our design builds on the MetaFormer architecture \cite{yu2022metaformer, yu2023metaformer}, which is a general vision transformer paradigm. In the Metaformer framework, the token mixer (e.g. self-attention, depthwise convolution) is not specified, while normalization, a channel mixer (e.g. MLP), and residual connections \cite{he2016deep} are retained. The input image tensor \( I \) is first passed through a patch embedding block, such as a convolution:
\begin{align}
X = \text{PatchEmbed}(I)=Flatten(\text{Conv}^{(k, s, p)}_{C_0 \rightarrow C} (I)) \label{eq1}
\end{align}
where \( X \in \mathbb{R}^{N \times C} \) denotes the embedded tokens with sequence length \( N \) and embedding dimension \( C \). Here, $C_0, k, s$ and $p$ represent input image channels, kernel size, stride and padding, respectively. These embedded tokens are then fed into the Metaformer architecture: 
\begin{align}
X' &= X + \text{TokenMixer}(\text{Norm}_1(X)), \label{eq2} \\
X'' &= X' + \sigma(\text{Norm}_2(X')W_1)W_2. \label{eq3}
\end{align}
Here, $\text{Norm}_1$ and $\text{Norm}_2$ are typically batch normalization \cite{ioffe2015batch} or layer normalization \cite{ba2016layer}. The TokenMixer serves as the core module for spatial information interaction, $W_1$ and $W_2$ are learnable weights in a two-layer channel MLP, and $\sigma$ is a non-linear activation \cite{nair2010rectified, hendrycks2016gaussian}. 

To further enhance the MetaFormer architecture, we introduce Decomposed Convolution as the token mixer within the DCFormer block. This design leverages the computational efficiency of decomposed 1D convolutional operations along each spatial axis (height, width, and depth). By splitting the 3D convolution into three parallel 1D convolutions, Decomposed Convolution captures spatial features while significantly reducing the number of parameters and computational cost. Thus, the DCFormer block integrates decomposed convolutions as a lightweight yet powerful token mixer. Let \( X \in \mathbb{R}^{B \times C \times H \times W \times D} \) denote the input feature map, where \( B \) is the batch size, \( C \) is the number of channels, and \( H \), \( W \), and \( D \) represent the spatial dimensions (height, width, and depth), respectively. The decomposed convolution consists of three 1D depthwise convolutions, processing the input tensor along each spatial axis:
\begin{align}
X^*_{\text{h}} &= \text{DWConv}^{k_h \times 1 \times 1}_{C \rightarrow C} (X), \\
X^*_{\text{w}} &= \text{DWConv}^{1 \times k_w \times 1}_{C \rightarrow C} (X), \\
X^*_{\text{d}} &= \text{DWConv}^{1 \times 1 \times k_d}_{C \rightarrow C} (X) 
\end{align}
where ($k_h, k_w, k_d$) represents the kernel sizes in height, width and depth dimensions, respectively. In our implementation, we set $k_h = k_w = k_d = k \in \{13, 11, 9, 7\}$ to leverage large kernels while maintaining computational efficiency through decomposition.
After applying decomposed convolutions along each spatial axis, we normalize the resulting features separately and then combine them using elementwise summation to form the DCFormer block: 
\begin{align}
X' &= X + \text{Norm}_h(X^*_{\text{h}}) + \text{Norm}_w(X^*_{\text{w}}) + \text{Norm}_d(X^*_{\text{d}})
\end{align}

An illustration of the DCFormer block and its relationship to existing architectures is shown in Figure \ref{fig:deconvnext}, and PyTorch-style pseudocode for the Decomposed Convolution operation is provided in Algorithm \ref{algo:decompconv3d}, respectively.

\input{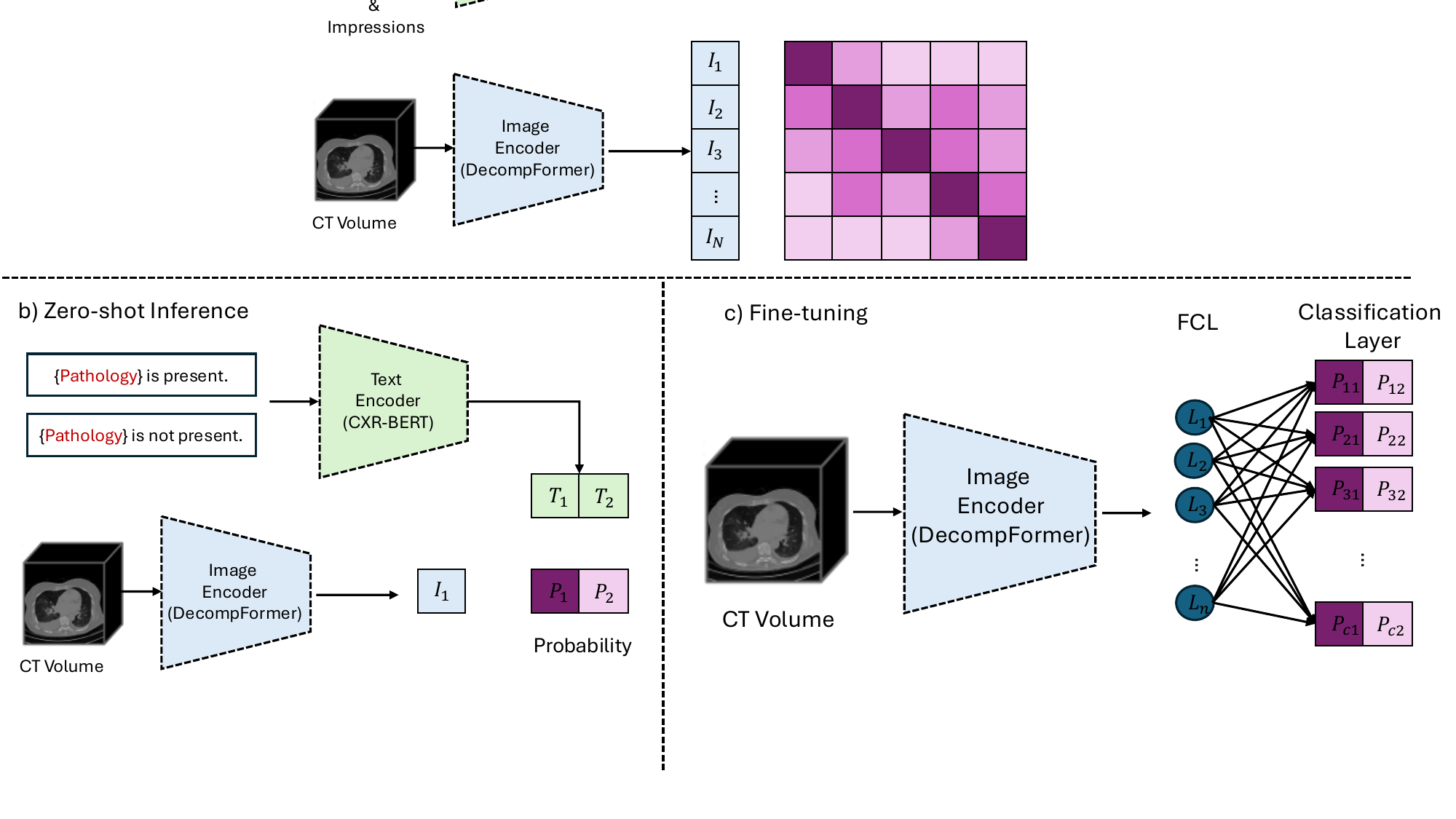}
\input{algorithm}

\subsection*{Complexity Analysis}
Given an input tensor \( X \in \mathbb{R}^{C \times H \times W \times D} \), a standard 3D depthwise convolution \cite{chollet2017xception} with a kernel size \( k \) in all dimensions has \( Ck^3 \) parameters and \( 2CHWDk^3 \) FLOPs (bias operations are omitted for simplicity). This cubic growth in complexity can quickly become prohibitive for large \( k \) and for deep networks, where multiple such layers compound the cost. In contrast, our decomposed 3D convolution uses three 1D depthwise convolutions, which collectively require only \( 3Ck \) parameters and \( 6CHWDk \) FLOPs. Thus, the parameter count and computational cost scale linearly with kernel size rather than cubically. Figure~\ref{fig:compdecomp} compares the theoretical FLOPs and parameter counts of a standard 3D depthwise convolution with those of a decomposed convolution, highlighting the drastic efficiency gain of the latter. Notably, for any kernel size larger than 3, the difference in complexity becomes increasingly significant. Large kernel sizes (e.g., 7 or 11, as used in prior works \cite{liu2022convnet, yu2024inceptionnext}, and even 51 as explored in recent studies \cite{ding2022scaling, liu2022more}) have been shown to improve CNN performance. In 3D imaging, our decomposition approach makes the use of such large kernels feasible, which is particularly important in deep networks where multiple convolutional layers are stacked.

\input{figures/overview}

\subsection*{Building a Vision Encoder with \dc}

Using the DCFormer block as the core component, we constructed a hierarchical vision encoder architecture (Figure \ref{fig:overview}). Following prior work \cite{he2016deep, liu2021swin, liu2022convnet, yu2022metaformer, wang2023internimage, yu2024inceptionnext}, DCFormer includes an initial stem stage followed by four hierarchical stages. In the stem stage, we use a decomposed convolution with a kernel size of 7 and a stride of 4. The downsampled images are then processed with three additional decomposed convolutions, each with a kernel size of 3 and a stride of 1. The output of the stem is then fed into four stages of DCFormer blocks, where each stage produces tokens of sizes \(\frac{H}{8} \times \frac{W}{8} \times \frac{D}{8}\), \(\frac{H}{16} \times \frac{W}{16} \times \frac{D}{16}\), \(\frac{H}{32} \times \frac{W}{32} \times \frac{D}{32}\), and \(\frac{H}{64} \times \frac{W}{64} \times \frac{D}{64}\), respectively, where $H$, $W$, $D$ represent height, width and depth of the original image volume. 

We implemented DCFormer in three model variants: nano, naive, and tiny. The number of layers in each stage is [1, 1, 1, 1] for nano, [2, 2, 2, 2] for naive, and [2, 3, 3, 2] for tiny. At the beginning of each stage (except the stem), we include a patch embedding downsampling operation implemented via 3D max pooling with a kernel size of 3 and a stride of 2, to further reduce the feature map size \cite{dai2021coatnet}. We set the MLP expansion ratio to 4 in all DCFormer blocks, following \cite{dosovitskiy2020image, liu2022convnet, yu2024inceptionnext}, and use a base decomposed convolution kernel size of 7, consistent with the design in \cite{liu2021swin, liu2022convnet}. Additional configuration details for each model variant are provided in Table~\ref{config}.
\input{tables/config}

\subsection*{Developing a CLIP Framework using \dc}
The CLIP framework aligns visual and contextual embeddings in a shared latent space through contrastive learning (Figure \ref{fig:clip}(a)). In this paper, we integrate our proposed DCFormer as the vision encoder in CLIP, serving as the primary module for efficient processing of 3D medical images. For the text encoder, we incorporate CXR-BERT \cite{boecking2022making}, following the implementation in Hamamci et al.\cite{hamamci2024foundation}. To compute similarities between images and radiology reports, we project visual and textual features into a shared 512-dimensional embedding space. Specifically, we apply global average pooling, followed by a linear projection to the output of the image encoder, while the text encoder output is transformed using a separate linear layer. Finally, visual and textual embeddings are aligned using a contrastive loss function \cite{radford2021learning}:
\begin{align}
    \mathcal{L}_{\text{CLIP}} = - \frac{1}{B} \sum_{i=1}^B \left[ \log \frac{\exp\left(\text{sim}(z_i^t, z_i^v)/\tau\right)}{\sum_{j=1}^B \exp\left(\text{sim}(z_i^t, z_j^v)/\tau\right)} + \log \frac{\exp\left(\text{sim}(z_i^v, z_i^t)/\tau\right)}{\sum_{j=1}^B \exp\left(\text{sim}(z_i^v, z_j^t)/\tau\right)} \right]
\end{align}
where $\text{sim}(z_i^t, z_j^v)$ represents the cosine similarity between the text embedding $z_i^t$ for the $i-$th radiology report and the visual embedding $z_j^v$ for the $j-th$ 3D image, $\tau$ is the temperature parameter (set to 1), and $B$ is the number of text-image pairs in a training batch. 

Unlike CT-CLIP \cite{hamamci2024foundation}, which lacks explicit multi-scale feature extraction, our DCFormer-based CLIP framework introduces a hierarchical structure that captures multi-scale features across different spatial resolutions. This design enhances representation learning and improves the alignment between image and text embeddings by preserving both global context and fine-grained details—crucial for accurate 3D CT image interpretation. Additionally, integrating DCFormer helps address the computational challenges associated with 3D CT imaging. By utilizing decomposed convolutions, our framework effectively reduces both the parameter count and computational overhead while maintaining robust feature extraction capabilities. As a result, the DCFormer-based CLIP framework is well-suited for large-scale 3D medical imaging applications.
\input{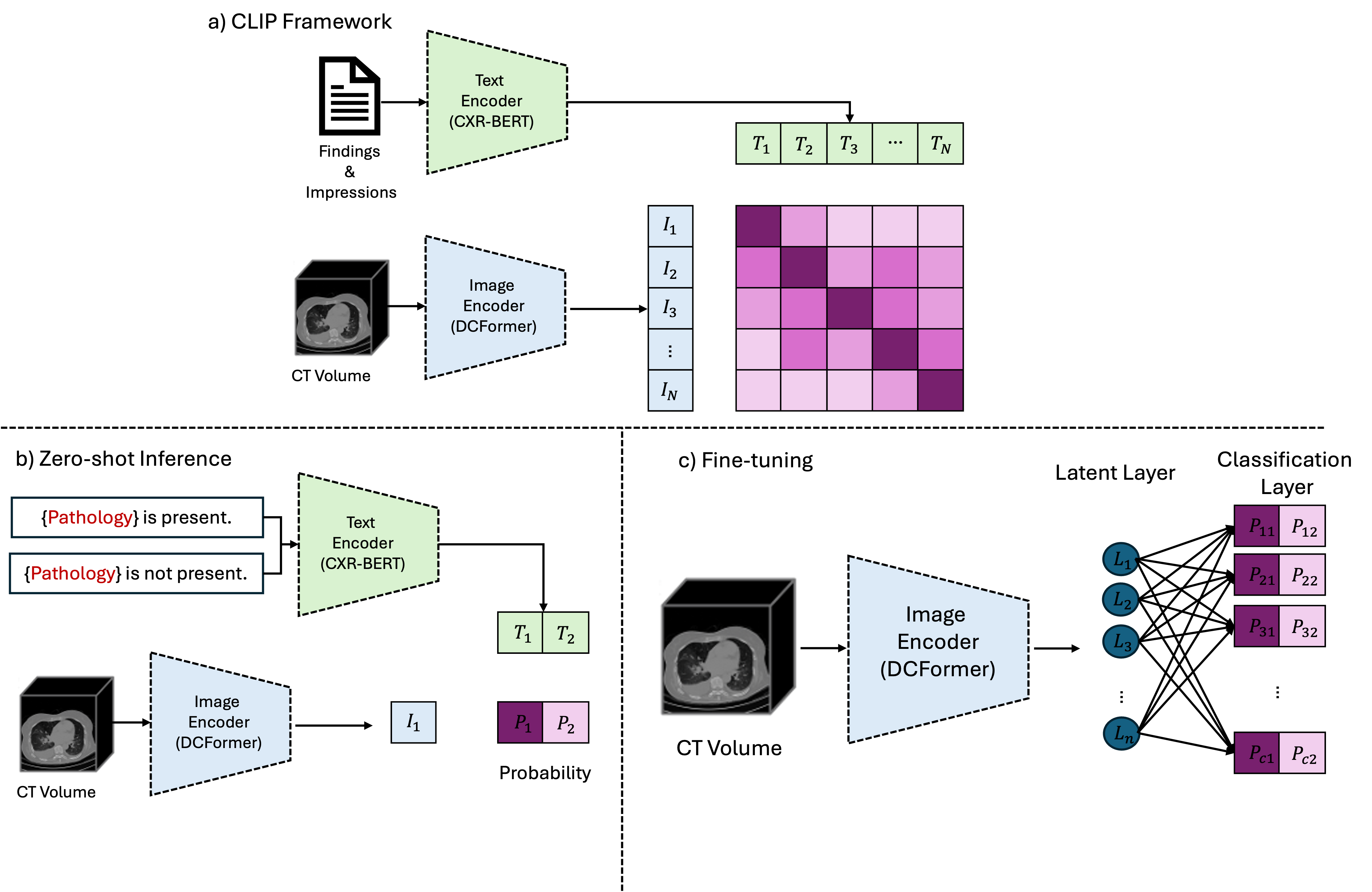}

\subsection*{State-of-the-Art Image Encoders}
The pioneering work ViT \cite{dosovitskiy2020image} introduced the concept of processing images as sequences of patches, leveraging the self-attention mechanism for global feature extraction. However, its quadratic complexity imposes a significant computational burden, particularly for 3D medical images. TransUNet \cite{chen2021transunet} was the first model to incorporate ViTs into medical image segmentation, employing a hybrid architecture that combines CNNs with transformers. ConvNeXt \cite{liu2022convnet} demonstrated that 7×7 depthwise convolutions can serve as an effective token mixer, improving performance over self-attention while maintaining computational efficiency. PoolFormer \cite{yu2022metaformer} introduced a more efficient feature extraction mechanism than ConvNeXt, replacing self-attention with a simple average pooling operation to minimize computational overhead while preserving strong performance.

\subsection*{Datasets}

For model training and evaluation, we used the open-source CT-RATE dataset \cite{hamamci2024foundation}, which consists of 50,188 reconstructed CT volumes from 25,692 distinct CT experiments involving 21,304 patients. Each CT volume is paired with a radiology report. The dataset also includes 18 distinct abnormalities extracted from medical reports of each CT scan (findings and impressions). Table \ref{ctrate} provides an in-depth overview of CT-RATE, detailing the distribution of abnormalities across training and validation subsets. Each abnormality is associated with the number of samples in both sets, along with their respective ratios within the dataset. These ratios represent the proportion of samples for each abnormality relative to the total dataset. Notably, the ratios for the training and validation sets remain nearly identical across all abnormalities, ensuring that the validation set accurately reflects the training distribution, making it reliable for evaluating model performance.

\subsection*{Implementation Details}

For model training, we followed the same data splitting and pre-processing strategy as in \cite{hamamci2024foundation}. Specifically, we used 20,000 patients for training and 1,304 for validation. For pre-processing, we first resized each CT volume to a spacing of 0.75 mm on the x- and y-axes, and 1.5 mm on the z-axis. The volumes were then center-cropped or padded to a fixed size of 512 × 512 × 256. Finally, we clipped the Hounsfield Unit (HU) values of each CT image volume to the range \([-1000, 1000]\) and normalized them to \([-1, 1]\). We trained our models using eight NVIDIA A100 80GB GPUs. For optimization, we used AdamW \cite{kingma2014adam} with an initial learning rate of \( 10^{-5} \). We did not apply learning rate scheduling or warmup, as we did not observe any significant improvement. All models were trained for 15 epochs. For fine-tuning, we integrated a linear classification layer into the image encoder and trained only this layer while keeping the image encoder frozen.

\input{tables/ctrate}

%% file: figures/deconvnext.tex
\begin{figure*}[t]
\centering
\includegraphics[width=11cm]{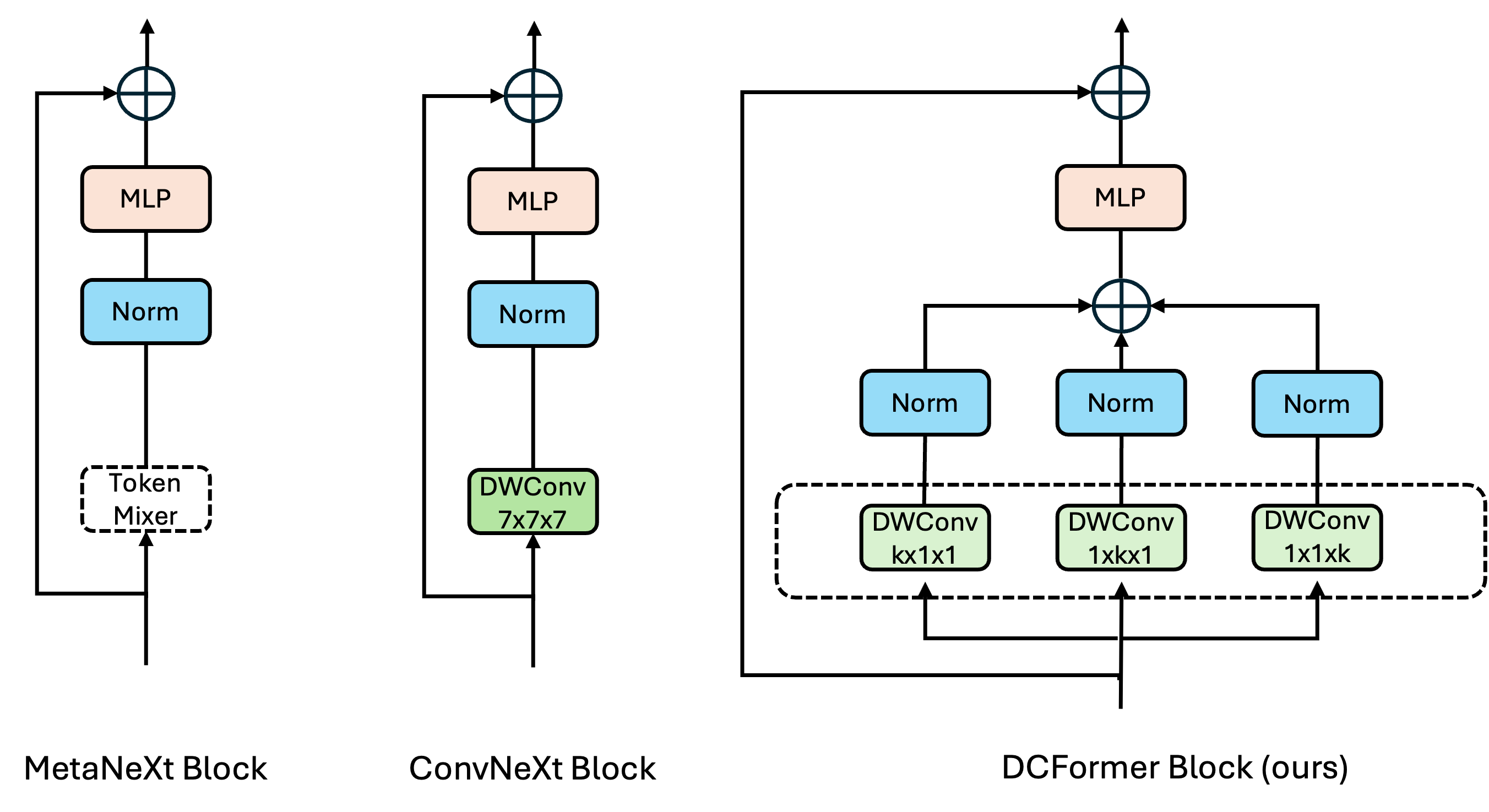}\\
\caption{Block illustration of MetaNeXt, ConvNeXt and \dc.}
\label{fig:deconvnext} 
\end{figure*}

%% file: algorithm.tex
\definecolor{commentcolor}{RGB}{110,154,155}   % Define comment color
\newcommand{\PyComment}[1]{\ttfamily\textcolor{commentcolor}{\# #1}}  % Add "#" before text
\newcommand{\PyCode}[1]{\ttfamily\textcolor{black}{#1}} % Python code font

\begin{algorithm}[h]
\SetAlgoLined
    \PyCode{\textcolor{magenta}{class} DecompConv3D(nn.Module):} \\
    \Indp
        \PyCode{\textcolor{magenta}{def} \_\_init\_\_(self, dim, kernel\_size, stride=1, norm=True, act=None):} \\
        \Indp
            \PyCode{\textcolor{magenta}{super}().\_\_init\_\_()} \\
            \PyComment{Initialize final activation layer} \\            
            \PyCode{self.act = act} \\
            \PyComment{Initialize normalization layer} \\
            \PyCode{self.norm = nn.ModuleList([nn.BatchNorm3d(dim) if norm else nn.Identity() for \_ in range(3)])} \\
            \PyComment{Initialize decomposed convolution layers} \\
            \PyCode{self.c1 = nn.Sequential(} \\
            \Indp
                \PyCode{nn.Conv3d(dim, dim, kernel\_size=(kernel\_size, 1, 1),} \\
                \PyCode{padding=(kernel\_size//2, 0, 0), stride=stride, groups=dim),} \\
                \PyCode{self.norm[0])} \\
            \Indm
            \PyCode{self.c2 = nn.Sequential(} \\
            \Indp
                \PyCode{nn.Conv3d(dim, dim, kernel\_size=(1, kernel\_size, 1),} \\
                \PyCode{padding=(0, kernel\_size//2, 0), stride=stride, groups=dim),} \\
                \PyCode{self.norm[1])} \\
            \Indm
            \PyCode{self.c3 = nn.Sequential(} \\
            \Indp
                \PyCode{nn.Conv3d(dim, dim, kernel\_size=(1, 1, kernel\_size),} \\
                \PyCode{padding=(0, 0, kernel\_size//2), stride=stride, groups=dim),} \\
                \PyCode{self.norm[2])} \\
            \Indm
        \Indm

        \PyCode{\textcolor{magenta}{def} forward(self, x):} \\
        \Indp
            \PyCode{x = self.c1(x) + self.c2(x) + self.c3(x)} \\
            \PyCode{if self.act is not None:} \\
            \Indp
                \PyCode{x = self.act(x)} \\
            \Indm
            \PyCode{\textcolor{magenta}{return} x} \\
        \Indm
    \Indm
\caption{PyTorch-style pseudocode for DecompConv3D}
\label{algo:decompconv3d}
\end{algorithm}

%% file: figures/overview.tex
\begin{figure*}[t]
\centering
\includegraphics[width=15.5cm]{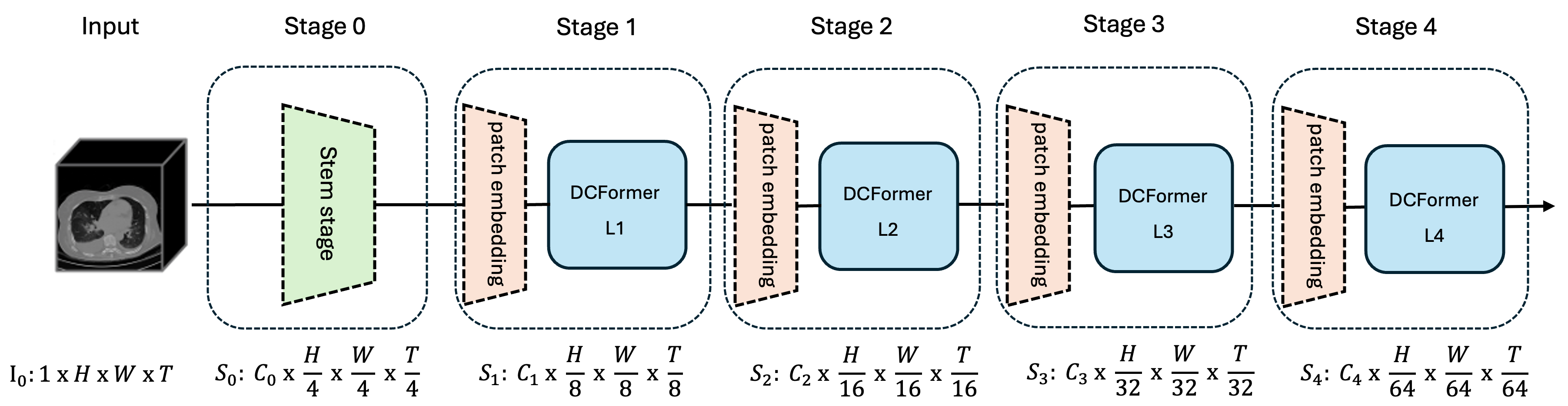}\\
\caption{Hierarchical architecture of DCFormer. }
\label{fig:overview} 
\end{figure*}

%% file: tables/config.tex
\begin{table*}[t]
\scriptsize
\renewcommand\arraystretch{1.4}
\begin{center}
\caption{\label{config} Architectural configurations of the DCFormer model variants.}
\begin{tabular}{llll|ccc}
\toprule
\multicolumn{1}{c|}{\multirow{2}{*}{Stage}} & \multicolumn{1}{c|}{\multirow{2}{*}{\#Tokens}}            & \multicolumn{2}{c|}{\multirow{2}{*}{Layer Specification}}                             & \multicolumn{3}{c}{DCFormer}                                                            \\ \cline{5-7} 
\multicolumn{1}{c|}{}                       & \multicolumn{1}{c|}{}                                    & \multicolumn{2}{l|}{}                                                                 & \multicolumn{1}{c|}{Nano}     & \multicolumn{1}{c|}{Naive}     & \multicolumn{1}{c}{Tiny}    \\ \midrule
\multicolumn{1}{l|}{\multirow{4}{*}{0}}     & \multicolumn{1}{l|}{\multirow{4}{*}{H/4 X W/4 X D/4}}    & \multicolumn{1}{l|}{\multirow{4}{*}{STEM}}               & Embed. Dim.                & \multicolumn{2}{c|}{32}       & \multicolumn{1}{c}{64}              \\ \cline{4-7} 
\multicolumn{1}{l|}{}                       & \multicolumn{1}{l|}{}                                    & \multicolumn{1}{l|}{}                                    & \#Blocks                    & \multicolumn{3}{c}{4}                                                                       \\ \cline{4-7} 
\multicolumn{1}{l|}{}                       & \multicolumn{1}{l|}{}                                    & \multicolumn{1}{l|}{}                                    & Patch size (block 1)       & \multicolumn{3}{c}{7 x 7 x 7, stride 4}                                                     \\ \cline{4-7} 
\multicolumn{1}{l|}{}                       & \multicolumn{1}{l|}{}                                    & \multicolumn{1}{l|}{}                                    & Patch size (block 2, 3, 4) & \multicolumn{3}{c}{3 x 3 x 3, stride 1}                                                     \\ \midrule
\multicolumn{1}{l|}{\multirow{5}{*}{1}}     & \multicolumn{1}{l|}{\multirow{5}{*}{H/8 X W/8 X D/8}}    & \multicolumn{1}{l|}{\multirow{2}{*}{Patch Embedding}}    & Patch size                 & \multicolumn{3}{c}{3 x 3 x 3, stride 2}                                                     \\ \cline{4-7} 
\multicolumn{1}{l|}{}                       & \multicolumn{1}{l|}{}                                    & \multicolumn{1}{l|}{}                                    & Embed. Dim.                & \multicolumn{1}{c|}{32}    & \multicolumn{1}{c|}{64}    & \multicolumn{1}{c}{96}            \\ \cline{3-7} 
\multicolumn{1}{l|}{}                       & \multicolumn{1}{l|}{}                                    & \multicolumn{1}{l|}{\multirow{3}{*}{DCFormer Block}} & Kernel size                & \multicolumn{3}{c}{7}                                                                       \\ \cline{4-7} 
\multicolumn{1}{l|}{}                       & \multicolumn{1}{l|}{}                                    & \multicolumn{1}{l|}{}                                    & MLP Ratio                  & \multicolumn{3}{c}{4}                                                                       \\ \cline{4-7} 
\multicolumn{1}{l|}{}                       & \multicolumn{1}{l|}{}                                    & \multicolumn{1}{l|}{}                                    & \#Blocks                    & \multicolumn{1}{c|}{1}     & \multicolumn{2}{c}{2}                                          \\ \midrule
\multicolumn{1}{l|}{\multirow{5}{*}{2}}     & \multicolumn{1}{l|}{\multirow{5}{*}{H/16 X W/16 X D/16}} & \multicolumn{1}{l|}{\multirow{2}{*}{Patch Embedding}}    & Patch size                 & \multicolumn{3}{c}{3 x 3 x 3, stride 2}                                                     \\ \cline{4-7} 
\multicolumn{1}{l|}{}                       & \multicolumn{1}{l|}{}                                    & \multicolumn{1}{l|}{}                                    & Embed. Dim.                & \multicolumn{1}{c|}{64}    & \multicolumn{1}{c|}{128}   & \multicolumn{1}{c}{192}           \\ \cline{3-7} 
\multicolumn{1}{l|}{}                       & \multicolumn{1}{l|}{}                                    & \multicolumn{1}{l|}{\multirow{3}{*}{DCFormer Block}} & Kernel size                & \multicolumn{3}{c}{7}                                                                       \\ \cline{4-7} 
\multicolumn{1}{l|}{}                       & \multicolumn{1}{l|}{}                                    & \multicolumn{1}{l|}{}                                    & MLP Ratio                  & \multicolumn{3}{c}{4}                                                                       \\ \cline{4-7} 
\multicolumn{1}{l|}{}                       & \multicolumn{1}{l|}{}                                    & \multicolumn{1}{l|}{}                                    & \#Blocks                    & \multicolumn{1}{c|}{1}     & \multicolumn{1}{c|}{2}     & \multicolumn{1}{c}{3}             \\ \midrule
\multicolumn{1}{l|}{\multirow{5}{*}{3}}     & \multicolumn{1}{l|}{\multirow{5}{*}{H/32 X W/32 X D/32}} & \multicolumn{1}{l|}{\multirow{2}{*}{Patch Embedding}}    & Patch size                 & \multicolumn{3}{c}{3 x 3 x 3, stride 2}                                                     \\ \cline{4-7} 
\multicolumn{1}{l|}{}                       & \multicolumn{1}{l|}{}                                    & \multicolumn{1}{l|}{}                                    & Embed. Dim.                & \multicolumn{1}{c|}{128}   & \multicolumn{1}{c|}{256}   & \multicolumn{1}{c}{384}           \\ \cline{3-7} 
\multicolumn{1}{l|}{}                       & \multicolumn{1}{l|}{}                                    & \multicolumn{1}{l|}{\multirow{3}{*}{DCFormer Block}} & Kernel size                & \multicolumn{3}{c}{7}                                                                       \\ \cline{4-7} 
\multicolumn{1}{l|}{}                       & \multicolumn{1}{l|}{}                                    & \multicolumn{1}{l|}{}                                    & MLP Ratio                  & \multicolumn{3}{c}{4}                                                                       \\ \cline{4-7} 
\multicolumn{1}{l|}{}                       & \multicolumn{1}{l|}{}                                    & \multicolumn{1}{l|}{}                                    & \#Blocks                    & \multicolumn{1}{c|}{1}     & \multicolumn{1}{c|}{2}     & \multicolumn{1}{c}{3}       \\ \midrule
\multicolumn{1}{l|}{\multirow{5}{*}{4}}     & \multicolumn{1}{l|}{\multirow{5}{*}{H/64 X W/64 X D/64}} & \multicolumn{1}{l|}{\multirow{2}{*}{Patch Embedding}}    & Patch size                 & \multicolumn{3}{c}{3 x 3 x 3, stride 2}                                                     \\ \cline{4-7} 
\multicolumn{1}{l|}{}                       & \multicolumn{1}{l|}{}                                    & \multicolumn{1}{l|}{}                                    & Embed. Dim.                & \multicolumn{1}{c|}{256}   & \multicolumn{1}{c|}{512}   & \multicolumn{1}{c}{768}           \\ \cline{3-7} 
\multicolumn{1}{l|}{}                       & \multicolumn{1}{l|}{}                                    & \multicolumn{1}{l|}{\multirow{3}{*}{DCFormer Block}} & Kernel size                & \multicolumn{3}{c}{7}                                                                       \\ \cline{4-7} 
\multicolumn{1}{l|}{}                       & \multicolumn{1}{l|}{}                                    & \multicolumn{1}{l|}{}                                    & MLP Ratio                  & \multicolumn{3}{c}{4}                                                                       \\ \cline{4-7} 
\multicolumn{1}{l|}{}                       & \multicolumn{1}{l|}{}                                    & \multicolumn{1}{l|}{}                                    & \#Blocks                    & \multicolumn{1}{c|}{1}     & \multicolumn{2}{c}{2}                                          \\ \midrule
\multicolumn{4}{c|}{Parameters (M)}                                                                                                                                                            & \multicolumn{1}{c|}{0.92} & \multicolumn{1}{c|}{5.85} & \multicolumn{1}{c}{15.1} \\ \midrule
\multicolumn{4}{c|}{FLOPS(G)}                                                                                                                                                                  & \multicolumn{1}{c|}{34.21} & \multicolumn{1}{c|}{49.48} & \multicolumn{1}{c}{168.2} \\
\bottomrule
\end{tabular}
\end{center}

\end{table*}

%% file: figures/clip.tex
\begin{figure*}[t]
\centering
\includegraphics[width=15.5cm]{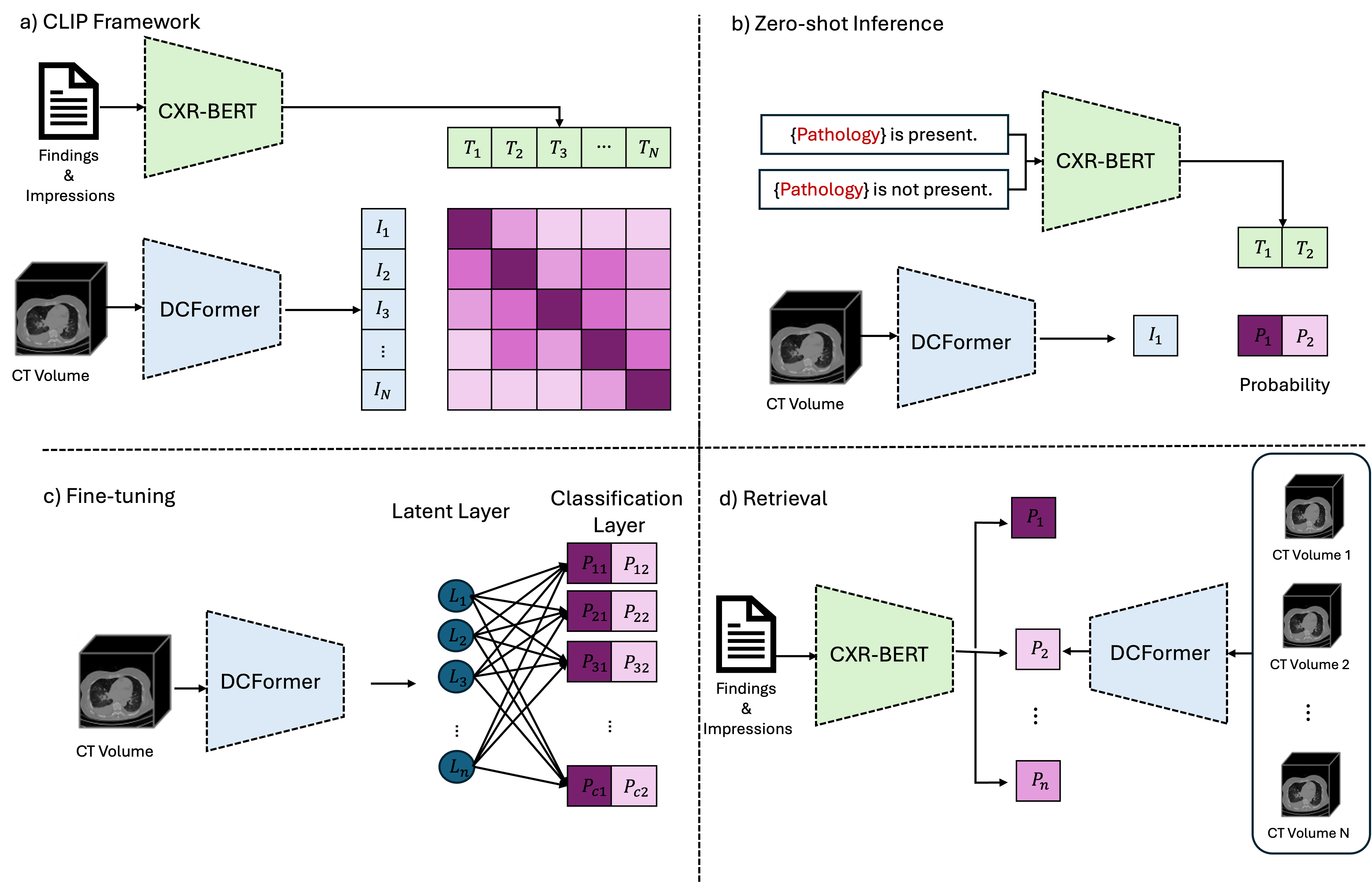}\\
\caption{DCFormer-based CLIP framework: (a) Training with paired CT volumes and reports, (b) Zero-shot inference with text prompts, (c) Fine-tuning for multi-label classification, and (d) Text-to-image retrieval based on embedding similarity.}
\label{fig:clip} 
\end{figure*}

%% file: tables/ctrate.tex
\begin{table*}[!th]
\scriptsize
\renewcommand\arraystretch{1.4}
\begin{center}
\caption{\label{ctrate} Detailed overview of the CT-RATE dataset for each abnormality and their distributions in the training and validation sets. \cite{hamamci2024foundation}}
\begin{tabular}{|l|c|c|c|c|}
\hline
Abnormality                        & \multicolumn{1}{l|}{Number of samples (train)} & \multicolumn{1}{l|}{Number of samples (validation)} & \multicolumn{1}{l|}{Ratio in the dataset (train)} & \multicolumn{1}{l|}{Ratio in the dataset (validation)} \\ \hline
Medical Material                   & 5818                                           & 313                                                 & 0.123                              & 0.103                                   \\ \hline
Arterial wall calcification        & 13377                                          & 867                                                 & 0.284                              & 0.285                                   \\ \hline
Cardiomegaly                       & 5308                                           & 325                                                 & 0.113                              & 0.107                                   \\ \hline
Pericardial effusion               & 3412                                           & 226                                                 & 0.072                              & 0.074                                   \\ \hline
Coronary artery wall calcification & 12025                                          & 765                                                 & 0.255                              & 0.252                                   \\ \hline
Hiatal hernia                      & 6751                                           & 417                                                 & 0.143                              & 0.137                                   \\ \hline
Lymphadenopathy                    & 12221                                          & 789                                                 & 0.259                              & 0.260                                   \\ \hline
Emphysema                          & 9122                                           & 600                                                 & 0.193                              & 0.197                                   \\ \hline
Atelectasis                        & 12263                                          & 713                                                 & 0.260                              & 0.235                                   \\ \hline
Lung nodule                        & 21382                                          & 1361                                                & 0.453                              & 0.448                                   \\ \hline
Lung opacity                       & 17420                                          & 1184                                                & 0.369                              & 0.390                                   \\ \hline
Pulmonary fibrotic sequela         & 12589                                          & 831                                                 & 0.267                              & 0.273                                   \\ \hline
Pleural effusion                   & 5705                                           & 376                                                 & 0.121                              & 0.124                                   \\ \hline
Mosaic attenuation pattern         & 3638                                           & 253                                                 & 0.077                              & 0.083                                   \\ \hline
Peribronchial thickening           & 4973                                           & 355                                                 & 0.105                              & 0.117                                   \\ \hline
Consolidation                      & 8319                                           & 581                                                 & 0.176                              & 0.191                                   \\ \hline
Bronchiectasis                     & 4732                                           & 330                                                 & 0.100                              & 0.109                                   \\ \hline
Interlobular septal thickening     & 3745                                           & 249                                                 & 0.079                              & 0.082                                   \\ \hline
\end{tabular}
\label{table1}
\end{center}

\end{table*}